\documentclass[runningheads]{llncs}

\usepackage{colortbl}

\usepackage{eccv}

\usepackage{eccvabbrv}
\definecolor{LightBlue}{cmyk}{0.09, 0.03, 0.01, 0.0}
\usepackage{graphicx}
\usepackage{booktabs}
\usepackage{wrapfig}
\usepackage[accsupp]{axessibility}  

\usepackage{hyperref}

\usepackage{orcidlink}

\graphicspath{{./figures/}}

\ifdefined \GramaCheck
\newcommand{\CheckRmv}[1]{}
\newcommand{\figref}[1]{Figure 1}%
\newcommand{\tabref}[1]{Table 1}%
\newcommand{\secref}[1]{Section 1}
\renewcommand{\eqref}[1]{Equation 1}
\else
\newcommand{\CheckRmv}[1]{#1}
\newcommand{\figref}[1]{Fig.~\ref{#1}}%
\newcommand{\tabref}[1]{Tab.~\ref{#1}}%
\newcommand{\secref}[1]{Sec.~\ref{#1}}
\renewcommand{\eqref}[1]{Eqn.~(\ref{#1})}

\fi

\usepackage{marvosym}

\usepackage{afterpage}
\usepackage{algorithm}
\usepackage{algpseudocode}
\usepackage{overpic}
\usepackage{tabularx} 
\usepackage{ragged2e} 
\usepackage{booktabs}
\usepackage[accsupp]{axessibility} 
\usepackage{multirow}
\usepackage{makecell}

\RequirePackage{fix-cm}

\newcommand{\minisection}[1]{ \noindent {\bf #1}\ \ }

\begin{document}

\title{Early Preparation Pays Off: New Classifier Pre-tuning for Class Incremental Semantic Segmentation
}

\titlerunning{Early Preparation Pays Off: New Classifier Pre-tuning for CISS}

\author{Zhengyuan Xie\inst{1} \orcidlink{0000-0003-2522-285X} \and
Haiquan Lu\inst{1} \orcidlink{0009-0003-2115-9782 } \and
Jia-wen Xiao\inst{1} \and
Enguang Wang\inst{1} \orcidlink{0009-0002-5161-913X} \and \\
Le Zhang \inst{3} \and
Xialei Liu \inst{1,2}\textsuperscript{(\Letter)}\orcidlink{0000-0001-8534-3026}}
\authorrunning{Z.~Xie et al.}
\institute{VCIP, CS, Nankai University \and
NKIARI, Shenzhen Futian \\
\email{\{xiezhengyuan\}@mail.nankai.edu.cn} \\
\email{\{xialei\}@nankai.edu.cn} \and
SICE, UESTC \\
 }

\maketitle

\begin{abstract}

Class incremental semantic segmentation aims to preserve old knowledge while learning new tasks, however, it is impeded by catastrophic forgetting and background shift issues.
Prior works indicate the pivotal importance of initializing new classifiers and mainly focus on transferring knowledge from the background classifier or preparing classifiers for future classes, 
neglecting the flexibility and variance of new classifiers.
In this paper, we propose a new classifier pre-tuning~(NeST) method applied before the formal training process, learning a transformation from old classifiers to generate new classifiers for initialization rather than directly tuning the parameters of new classifiers.
Our method can make new classifiers align with the backbone and adapt to the new data, preventing drastic changes in the feature extractor when learning new classes.
Besides, we design a strategy considering the cross-task class similarity to initialize matrices used in the transformation, helping achieve the stability-plasticity trade-off.
Experiments on Pascal VOC 2012 and ADE20K datasets show that the proposed strategy can significantly improve the performance of previous methods. The code is available at \url{https://github.com/zhengyuan-xie/ECCV24_NeST}.

\end{abstract}

\section{Introduction}\label{sec:intro}

Nowadays, segmentation models driven by deep neural networks have achieved notable success in various fields~\cite{long2015fully, chen2017no, chen2018encoder}.
However, these models are usually trained and tested in a static environment~\cite{geng2020recent}, which is contrary to real-world scenarios~\cite{wang2024unlockingmultimodalpotentialclip}. 
When facing a continuous data stream~\cite{hadsell2020embracing}, a model needs to handle several novel classes.
Blending new and previous data to retrain the model, or called \emph{Joint Training}, ensures the performance of the model.
However, the computational cost becomes prohibitive when dealing with large datasets.
Simply fine-tuning the model may lead to the loss of previously acquired knowledge, a phenomenon known as catastrophic forgetting~\cite{mccloskey1989catastrophic, kirkpatrick2017overcoming}. 
In particular, for semantic segmentation, it also faces the problem of background shift~\cite{mib}~(i.e., pixels labeled as background in the current step may belong to previous or future classes). 

\begin{figure}[t]
	\small
	\centering
	\includegraphics[width=0.99\linewidth]{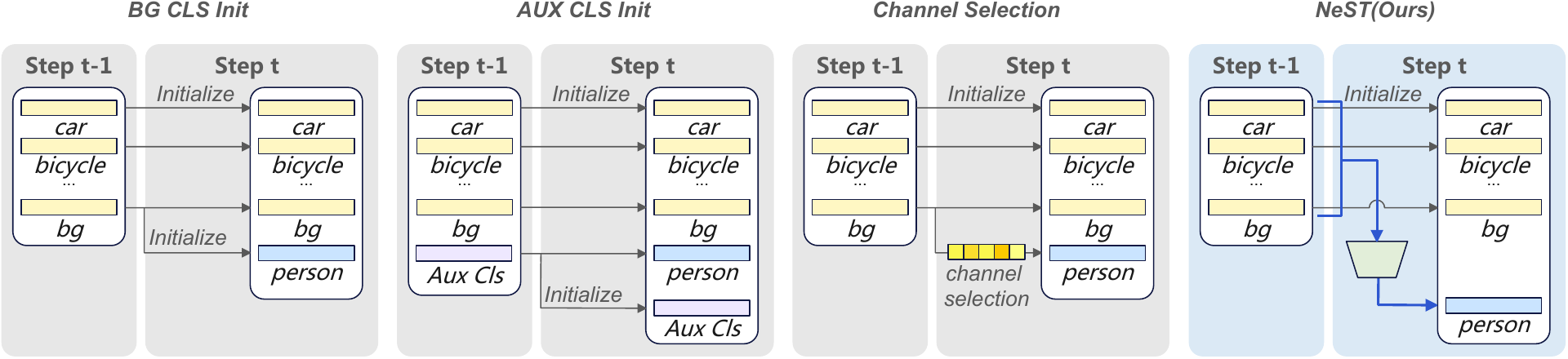}
	\caption{ 
    Different classifier initialization methods for class incremental semantic segmentation. MiB~\cite{mib} directly uses background classifiers to initialize new classifiers. Some methods~\cite{ssul, dkd} train an auxiliary classifier for future classes. AWT~\cite{awt} selects the most relevant weights from the background classifier for new classifiers' initialization by gradient-based attribution. Our new classifier pre-tuning method learns a transformation from all old classifiers to generate new classifiers for initialization.
	}\label{fig:comparison}
\end{figure}

Class incremental semantic segmentation (CISS)~\cite{mib,plop,rcil,ssul} has been proposed to address the challenges of catastrophic forgetting and background shift.
It requires the segmentation model to learn concepts of new classes while preserving old knowledge.
If no restrictions are imposed, parameters crucial to old classes may be updated in the wrong direction~\cite{kim2023stability}, causing the model to forget previously learned knowledge.
On the contrary, only focusing on memorizing old knowledge may limit the model to learning new classes.
Thus, the key issue to continual learning is actually the trade-off between stability and plasticity~\cite{grossberg2012studies, kim2023stability}. 

As far as stability is concerned, existing methods rely on old models learned from the previous step for knowledge transfer~\cite{mib,plop,rcil,ewf}, \eg  \emph{weight initialization}. 
Yet the old model does not have the ability to recognize new classes, finding a proper way to initialize the newly added classifier is crucial.
Random initialization may cause the misalignment between new classifiers and features, leading to training instabilities~\cite{mib}. 
In accordance with this situation, as~\cref{fig:comparison} shows, MiB~\cite{mib} utilizes the background classifier to initialize new classifiers, as future classes may appear in current data, labeled as the background.
However, it may cause the model to make incorrect predictions for true background pixels~\cite{awt}.
AWT~\cite{awt} selects the most relevant weight for initialization from the old background classifier via a gradient-based attribution technique, yet it neglects other old classifiers and brings a huge memory cost.
Some methods train an auxiliary classifier to initialize new classifiers~\cite{ssul, dkd}, while there is still a bias between the auxiliary classifier and true future classifiers because there is not any future data with ground truth.
What's more, the above initialization methods treat each new classifier equally, ignoring the differences between new classes.

From the above observations, we propose \underline{\textbf{N}}\underline{\textbf{e}}w cla\underline{\textbf{S}}sifier pre-\underline{\textbf{T}}uning~(NeST) method to make new classifiers better align with the backbone and adapt to the training data, preventing drastic changes in the feature extractor caused by unstable training process.
To better utilize the old knowledge, instead of directly tuning the parameters of new classifiers, we tune a linear transformation from all old classifiers to generate new classifiers.
Specifically, before the formal training process at the current step,
we assign two transformation matrices, \ie, {an importance matrix, and a projection matrix}, to each new class. 
As trainable parameters, the importance matrix learns a weighted score associated with each channel of the old classifier pertinent to the new class, while the projection matrix learns a linear combination from the weighted old classifiers to generate the new classifier.
Subsequently, we use data from the current step to learn the transformation from old classifiers to each new classifier. 
Finally, we employ learned importance matrices and projection matrices to generate new classifiers from old ones, and then leverage the derived weights for classifier initialization of the formal training process.
By using these two matrices, new classifiers are generated via old knowledge and are different from each other.

To achieve the stability-plasticity trade-off, we also find it crucial for the initial values of these transformation matrices and propose a strategy for initialization by considering cross-task class similarities between old and new classes, thus facilitating the learning of new classifiers.

To summarize, the main contributions of our paper are threefold:

\begin{enumerate}

    \item We propose a new classifier pre-tuning~(NeST) method that learns a transformation from all old classifiers to generate new classifiers with previous knowledge before the formal training process.
    
    \item We further optimize the initialization of transformation matrices, striking a balance between stability and plasticity.
    \item We conduct experiments on Pascal VOC 2012 and ADE20K. 
    Results show that NeST can be readily integrated into other approaches and significantly enhance their performance.
\end{enumerate}

\section{Related Work}\label{sec:related}
\minisection{Semantic Segmentation.}
In recent years, the development of deep learning techniques has facilitated the performance of semantic segmentation models.
Fully Convolutional Networks~(FCN)~\cite{fcn} pioneers semantic segmentation, and a series of convolution-based segmentation networks have achieved high performance in many benchmarks~\cite{unet, deeplabv3, zhang2018exfuse, wang2020deep,guo2022segnext, gao2019res2net}.
More recently, the Transformer architecture has become ever more popular and reached remarkable performance~\cite{segformer,maskformer,mask2former, ranftl2021vision, setr, segmenter, swin}.
Thus, we conduct experiments on two different backbones: ResNet and Swin-Transformer.

\minisection{Class Incremental Learning.}
Class incremental learning tackles the catastrophic forgetting of the task with ever-increasing categories.
The whole training process is divided into several steps and in each step, the model is required to learn one or more classes, which is the most challenging.
Existing methods can be categorized into three groups.
The \emph{model expansion} methods~\cite{yoon2017lifelong, hung2019compacting,hu2023dense, der} enlarge the model size in incremental steps to learn new knowledge while preserving old knowledge in fixed parameters.
The \emph{rehearsal based} methods store a series of exemplars~\cite{bang2021rainbow, verwimp2021rehearsal} or prototypes along the task sequence~\cite{zhu2021prototype}, or using generative networks to maintain old knowledge~\cite{shin2017continual, recall}.
The \emph{parameter regularization} methods either constrain the learning of some parameters~\cite{lwf,chaudhry2018riemannian} or use knowledge distillation techniques~\cite{podnet, wu2019large, aljundi2018memory, plop}.

\minisection{Class Incremental Semantic Segmentation.}
CISS requires a segmentation model to recognize all learned classes in continuous learning steps.
Different from classification tasks, segmentation has its own background shift~\cite{mib} issue.
MiB~\cite{mib} first proposes unbiased cross entropy and distillation to address background shift.
PLOP~\cite{plop} uses pseudo-label and intermediate features to transfer old knowledge.
SDR~\cite{sdr} proposes a contrastive-learning-based method, minimizing intra-class feature distances.
RCIL~\cite{rcil} introduces reparameterization to continual semantic segmentation with a complementary network structure.
GSC~\cite{cong2023gradient} explores incremental semantic segmentation considering the gradient and semantic compensation.
EWF~\cite{ewf} fuses old and new knowledge with a weight fusion strategy.
Many methods have explored the classifier initialization in CISS.
SSUL~\cite{ssul} employs auxiliary data to train an `unknown' classifier and use it to initialize new classifiers.
DKD~\cite{dkd} proposes a decomposed knowledge distillation, improving the rigidity and stability.
AWT~\cite{awt} applies gradient-based attribution to transfer the relevant weight of old classifiers to new classifiers.
Different from the method above, we propose a new classifier pre-tuning method, achieving the goal of the trade-off between plasticity and stability.

\section{Method}
\subsection{Preliminaries} \label{sec:prelimi}

\minisection{Problem Definition.} Following \cite{mib, plop}, CISS contains several steps $\{t\}_{t=1}^n$ to receive sequential data stream $\{\mathcal{D}_t\}_{t=1}^n$ with classes $\{\mathcal{C}_t\}_{t=1}^n$. 
In each step $t$, the model is required to learn $\left |\mathcal{C}_t \right |$ new classes, while old training data is not available.
For an image in the current step, pixels belonging to $\mathcal{C}_t$ are labeled as their ground truth classes, leaving other pixels labeled as background.
Finally, after the last step, the model will be tested on the data of all learned classes.

At step $t$, the segmentation model is composed of a feature extractor $f_{\theta}^t$ and a classifier $h_{\phi}^t$. 
$\phi_t$ is newly added and weights of previous classifiers are $\phi_{1:t-1}$.
In this paper, our main concern is the initialization of $\phi_t$.
\begin{figure*}[!htp]
	\small
	\centering
	\includegraphics[width=0.99\linewidth]{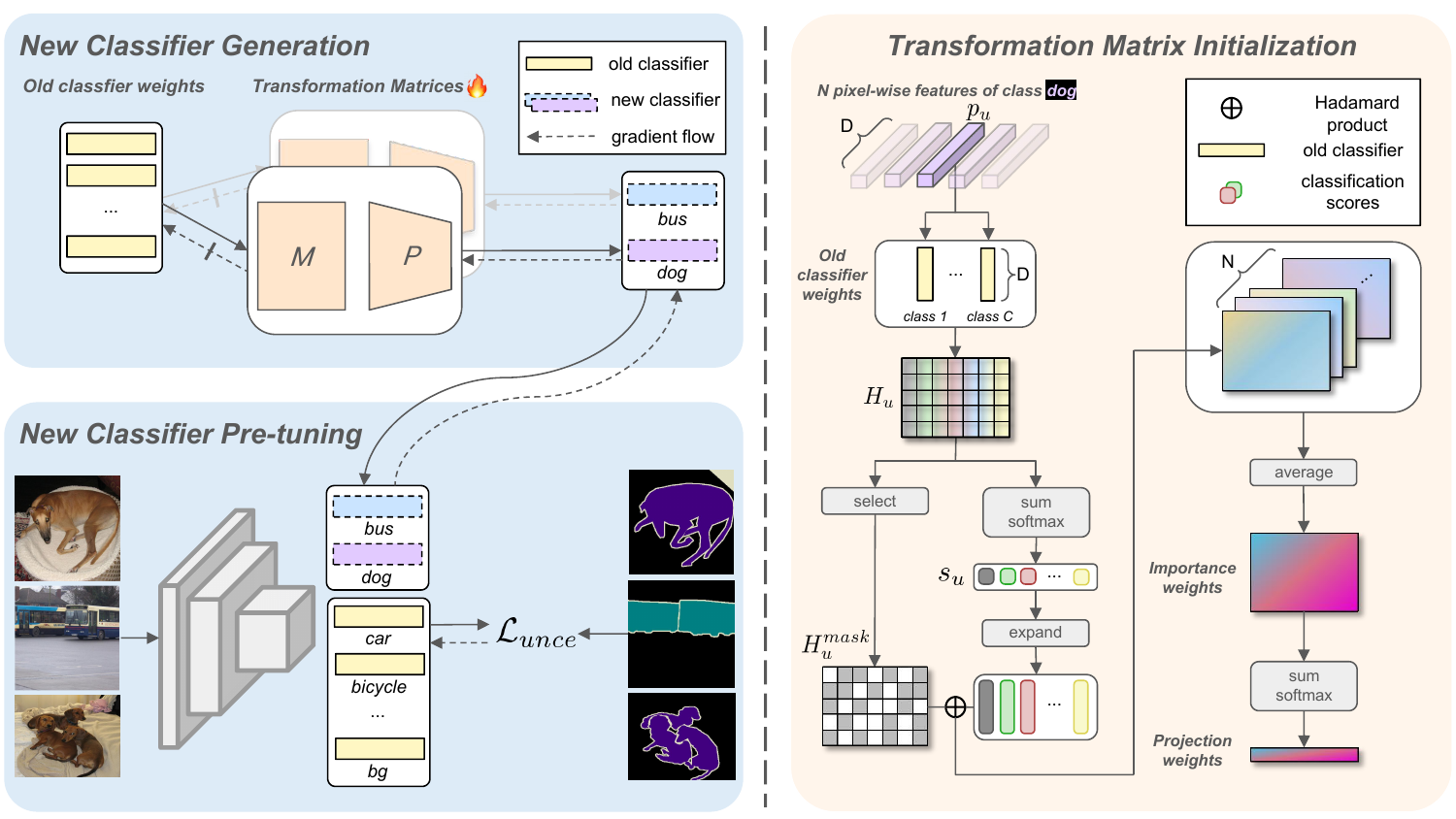}
	\caption{ 
Illustration of our new classifier pre-tuning~(NeST) method. The left side of the figure is an iteration of the new classifier pre-tuning process. The right side of the figure represents the cross-task class similarity-based initialization of importance matrices and projection matrices before the pre-tuning process.
	}\label{fig:framework}
\end{figure*}

\minisection{Revisiting Previous Initialization Methods.}
In this part, we illustrate the shortcomings of previous initialization methods and further clarify our motivation. 
As mentioned before, classes in $\mathcal{C}_{t}$ will appear as background in previous steps, which is known as the \emph{background shift}. 
According to this premise, previous methods either try to find a better way of utilizing the background classifier~\cite{mib, awt}, or train an auxiliary classifier for future classes, both neglecting the differences between new classes~\cite{ssul,dkd}. 
Meanwhile, there will also be a mismatching issue between new classifiers and the feature extractor as the methods above do not have training processes with data from new classes. This may lead to drastic parameter changes in the feature extractor when training with new data, thus undermining previous knowledge.

\subsection{New Classifier Pre-Tuning}
\label{sec:pre_tuning}

Considering the observations above, we propose a simple yet effective method, NeST, to improve the performance of CISS. 
In the pre-tuning stage before the formal training process, we first learn to generate each new classifier via old classifiers and then tune it with new data in each forward pass.

\minisection{New Classifier Generation.} As shown in~\cref{fig:framework}, for each new class $c\in \mathcal{C}_t$, we assign an importance matrix and a projection matrix, learning the transformation from all old classifiers in each forward pass as follows:
\begin{equation}
  \begin{aligned}
    \label{eqn:transform}
    \mathbf{w}_{c} &= (\mathbf{M}_c\odot \mathbf{W}_{old})\mathbf{P}_c\,,
  \end{aligned}
\end{equation}
where $\mathbf{w}_c \in \mathbb{R}^{d}$ denotes the weight of new class $c$, $\mathbf{M}_c\in \mathbb{R}^{d\times n_{old}}$ denotes the importance matrix of class $c$ and $n_{old}=\left | \bigcup_{i=1}^{t-1} \mathcal{C}_i\right | + 1$ denotes the number of old classes including the background, $\odot$ denotes the operation of Hadamard product, $\mathbf{W}_{old}\in \mathbb{R}^{d\times n_{old}}$ denotes the weight matrix of all old classifiers, $\mathbf{P}_c \in \mathbb{R}^{n_{old}\times 1}$ denotes the projection matrix of class $c$, and $d$ denotes the output dimension of $f_{\theta}^{t-1}$. 
$\mathbf{M}_c$ reflects the importance of each channel of the old classifiers for the new class $c$, and $\mathbf{P}_c$ combines weighted old classifiers, completing weight generation for the new class.
After generating new weights, we concatenate them as the parameter of $\phi_t$ as follows:
\begin{equation}
    \begin{aligned}
        \phi_t=\rm{concat} [\mathbf{w}_{n_{old}}, ..., \mathbf{w}_{n_{old}+\left |\mathcal{C}_{t}\right | - 1}]\,,
    \end{aligned}
\end{equation}
where $\text{concat}[   ]$ is the concatenation operation. 
Specifically, as the old background class may see new classes in previous steps and the score of background in the pre-tuning process may be high, preventing the model from learning new classes, we also learn a transformation from the old background classifier to the new background classifier during the pre-tuning process as follows:
\begin{equation}
  \begin{aligned}
    \label{eqn:transform_bg}
    \hat{\mathbf{w}}_{0} &= (\mathbf{M}_0\odot \mathbf{w}_{0})\mathbf{P}_0\,,
  \end{aligned}
\end{equation}
where $\mathbf{w}_{0}\in \mathbb{R}^{d}$ denotes the previous background classifier, $\mathbf{M}_0\in \mathbb{R}^{d\times 1}$ and $\mathbf{P}_0\in \mathbb{R}^{1\times 1}$ are matrices specifically for the background, and $\hat{\mathbf{w}}_{0}\in \mathbb{R}^{d}$ is new background classifier prepared for current task.  

\minisection{Pre-tuning and Initialization.}
In each forward pass, as~\cref{fig:framework} shows, after generating weights of new classifiers, we feed training data from the current step to the old model equipped with generated classifiers, using the output to calculate the loss and then backpropagating the loss to update learnable parameters.
The loss function used in the pre-tuning is unbiased cross entropy $\mathcal{L}_{unce}$~\cite{mib}, as it can help to avoid the overfitting problem~\cite{alife}:
\begin{equation}
\begin{aligned}
&\mathcal{L}_{unce}=-\frac{1}{\left| \mathcal{I} \right|} \sum_{i\in\mathcal{I}}\log \tilde{q}_x^t(i, y_i)\,,
\end{aligned}
    \label{eqn:unce}
\end{equation}
where $\mathcal{I}$ denotes the pixel set of an image, $y_i$ denotes the ground-truth label of pixel $i$ at current step, 
$\tilde{q}_x^t$ denotes the modified output of the current model.
It should be noted that during the whole pre-tuning process, the importance matrices and projection matrices are learnable and will be updated, while other components such as the feature extractor remain frozen. After the pre-tuning process, we use importance matrices and projection matrices to generate weights of each new classifier for initialization, and additional parameters will be removed.

\subsection{Cross-Task Class Similarity-based Transformation Matrix Initialization}
\label{sec:matrix_initialization}
We found it crucial to initialize importance matrices and projection matrices. 
Random initialization disregards the importance of different channels among old classifiers and results in poor performance.
AWT~\cite{awt} reveals that only a few channels of the background classifier contribute to the classification of new classes, redundant channels may hinder the learning of new classifiers. 
Meanwhile, for different new classes, the important channels of old classifiers should be different.
Hence, we introduce a cross-task class similarity-based transformation matrix initialization method. 
The core idea is that if an old class is more similar to a new class, 
then during the pre-tuning process,
the contribution of this old classifier should be more significant, and thus, greater initial weights should be assigned to corresponding positions in the importance matrix and projection matrix.

Hence, we employ the predictions made by the old model as cross-task class similarity scores for each new class pixel, aiming to assess the resemblance between the new class and every old class. 
As shown in \cref{fig:framework}, before learning transformation, we feed each training image in the current step to the old model and get the corresponding output of the layer before $h_{\phi}^{t-1}$.
Then considering the classification process of a new class pixel embedding $\mathbf{p}_u\in \mathbb{R}^{d}$ extracted from the old model,
assuming that the total number of old classes is $n_{old}$,
the matrix multiplication can be decomposed into an element-wise multiplication~(also called Hadamard product) between $\mathbf{p}_u$ and the old classifier weight $\mathbf{W}_{old}\in \mathbb{R}^{d\times n_{old}}$ and a sum-softmax operation, as follows:
\begin{equation}
\begin{split}
\mathbf{H}_{u} &= (\mathbf{W}_{old}\odot \mathbf{p}_u'),\\
\mathbf{s}_u &= {\rm softmax}({\rm sum}(\mathbf{H}_{u}))^{\top}\,,
\end{split}
\label{cross_task_similarity}
\end{equation}
where $\mathbf{p}_u'\in \mathbb{R}^{d\times n_{old}}$ denotes broadcasting $\mathbf{p}_u$ by $n_{old}$ times, $\mathbf{H}_u\in \mathbb{R}^{d\times n_{old}}$ denotes the result of the Hadamard product, ${\rm sum}$ denotes summing $\mathbf{H}_u$ along the channel dimension, and $\mathbf{s}_u\in \mathbb{R}^{n_{old}}$ denotes the classification scores of $\mathbf{p}_u$. 
We consider that for each column~(corresponding to an old class $i$) of $\mathbf{H}_u$, the positive elements contribute to $\mathbf{p}_u$ being classified into class $i$.
Thus, to select relevant channels, we set all positions with positive values to $1$ and get $\mathbf{H}_u^{mask}$, which can be regarded as a binary mask, as follows:
\begin{equation}
\mathbf{H}_u^{mask}(i, j)=
\left\{
\begin{aligned}
&1, \mathbf{H}_u(i, j)>0 \\
&0, otherwise\,.
\end{aligned}
\right.
    \label{H_mask}
\end{equation}

Finally, by utilizing ground-truth masks of the current step, for each pixel embedding belonging to new class $\boldsymbol{c_{new}}$, we calculate the Hadamard product between its corresponding $\mathbf{H}_{mask}$ and predicted score, then averaged the results to get the importance matrix weight $\mathbf{M}_{c_{new}}$ of class $c_{new}$, as follows:
\begin{equation}
    \begin{aligned}
        \mathbf{M}_{c_{new}} = \frac{1}{N} \sum_{\mathbf{p}_u} \mathbf{H}_u^{mask} \odot \mathbf{s}_u' \,,
    \end{aligned}
    \label{eqn:H_avg}
\end{equation}
where $\mathbf{p}_u$ denotes the pixel embedding belonging to the new class $c_{new}$, $\mathbf{s}_u'\in \mathbb{R}^{d\times n_{old}}$ denotes broadcasting the predicted score $\mathbf{s}_u$ of $\mathbf{p}_u$ by $d$ times, $N$ denotes the number of pixels belonging to new class $c_{new}$. Here we use $\mathbf{s}_u'$ because the old class with a very small score will not contribute too much to the initialization of the new class, as the similarity between these two classes is relatively low.
For projection matrix weight $\mathbf{P}_{c_{new}}$, we sum up $\mathbf{M}_{c_{new}}$ along the channel dimension and apply the softmax function to get the weight score for old classes, as follows:
\begin{equation}
    \begin{aligned}
        \mathbf{P}_{c_{new}}  = ({\rm softmax}({\rm sum}(\mathbf{M}_{c_{new}})))^{\top}\,.
    \end{aligned}
\end{equation}
Then we use $\mathbf{P}_{c_{new}}\in \mathbb{R}^{n_{old}\times 1}$ to initialize the projection matrix of class $c_{new}$ as these scores reflect the degree of similarity between the new class and old classes, deciding which old class the model should focus on for knowledge transfer.
We provide the pseudo-code of our NeST in~\cref{alg:process}.

\begin{algorithm}[!t]
	\renewcommand{\algorithmicrequire}{\textbf{Input:}}
	\renewcommand{\algorithmicensure}{\textbf{Output:}}
	\caption{Pseudo-code of the proposed NeST.} 
	\label{alg:process} 
	\begin{algorithmic}[1] 
		\Require 
		Training samples $\mathcal{D}_t = \{(x_i, y_i)\}_t$ of current task $t$, feature extractor $f_{\theta}^0$, classifier $h_{\phi}^{0}$, task number $T$, new class set $\mathcal{C}_t$.
		\Ensure 
		initial weight of new classifier $\phi_{t}$

		\For {$t\in$ $\{1,2,...,T\}$}
        
        \For {$i\in$ $\mathcal{C}_t$}
        \Comment{Initializing Transformation matrices. \cref{sec:matrix_initialization}}
        \State $(\mathbf{M}_i, \mathbf{P}_i)$ ← $Init(\mathcal{D}_t, h_{\phi}^{t-1}, f_{\theta}^{t-1})$
        \EndFor 
        \While {not converged}
        \Comment{Pre-tuning new classifiers. \cref{sec:pre_tuning}}
        \State train $(\mathbf{M}, \mathbf{P})$ by minimizing $\mathcal{L}_{unce}$
        
        \EndWhile
        \State $\phi_{t}$ ← ${\rm Transform}(\mathbf{M}, \mathbf{P}, \phi_{1:t-1})$
        \Comment{Initializing new classifiers.}
		\State $f_{\theta}^t$ ← $f_{\theta}^{t-1}$
        \State $h_{\phi}^t$ ← $(h_{\phi}^{t-1}, \phi_{t})$
		\While {not converged} 
        \Comment{Formal training process.}
        \State train $f_{\theta}^t$ and $h_{\phi}^t$ 
        
		\EndWhile
		\EndFor
	\end{algorithmic} 
\end{algorithm}

\section{Experiments}\label{sec:experiments}

\subsection{Experimental setup}\label{sec:setup}

\minisection{Protocols.}
In CISS, the whole training process contains $T$ steps, and the task of each step may contain one or more classes.
In step $t$, pixels belonging to previous steps are labeled as \emph{background}.
In evaluation, the model needs to identify all learned classes.
There are two settings: \emph{Disjoint} and \emph{Overlapped}~\cite{mib}.
\emph{Disjoint} setting assumes that in the current step, images do not contain any pixel belonging to classes that will be learned in the future. 
\emph{Overlapped} setting is more realistic, as future classes may appear in images from the current step, 
and we mainly evaluate our method on \emph{Overlapped} setting.

\minisection{Datasets.}
Pascal VOC 2012 dataset~\cite{pascal-voc-2012} contains 20 classes including 10,582 images for training and 1449 images for validation.
ADE20K dataset~\cite{ade} contains 150 classes including 20,210 images for training and 2000 images for validation.
In CISS, the training setting can be described as $X-Y$, $X$ means the number of classes trained in the initial step and $Y$ means the number of classes trained in incremental steps.
We conduct experiments with \emph{10-1}, \emph{15-1}, \emph{15-5} and \emph{19-1} settings on Pascal VOC 2012 dataset~\cite{pascal-voc-2012}.
For ADE20K dataset~\cite{ade}, we validate our method on \emph{100-50}, \emph{100-10}, \emph{100-5} and \emph{50-50} settings.

\minisection{Implementation Details.}
Following previous methods~\cite{mib, plop, rcil}, we use DeepLabV3~\cite{deeplabv3} with ResNet-101~\cite{resnet} as our segmentation model.
We also use Swin-B~\cite{swin} as the Transformer-based backbone.
Following \cite{mib, plop}, we use random crop and horizontal flip for data augmentation.
We train the model with a batch size of 24 on 4 GPUs for all experiments. 
We use SGD as the optimizer in our experiments.
The learning rate is set to 0.02 for the initial step and 0.001 for incremental steps, both for Pascal VOC 2012 and ADE20K.  
We also use \emph{poly} schedule as the weight decay strategy.
For Pascal VOC 2012, we pre-tune for 5 epochs with a learning rate of 0.001. For ADE20K, pre-tuning lasts 15 epochs, with a learning rate of 0.1 for experiments involving RCIL and Swin-B, and 0.5 for other experiments.
Further details are provided in the supplementary material.

\begin{table}[!t]\label{voc}
\centering
    \caption{The mIoU~(\%) of the last step on the Pascal VOC dataset for four different CISS scenarios. *~implies results from the re-implementation of the official code.}\label{tab:voc}
    \resizebox{1.0\textwidth}{!}{
    \begin{tabular}{c|c|ccc|ccc|ccc|ccc}
        \toprule
        \multirow{2}{*}{\textbf{Method}} & \multirow{2}{*}{\textbf{Backbone}} & \multicolumn{3}{c|}{\textbf{10-1(11 steps)}} & \multicolumn{3}{c|}{\textbf{15-1(6 steps)}} & \multicolumn{3}{c|}{\textbf{15-5(2 steps)}} & \multicolumn{3}{c}{\textbf{19-1(2 steps)}} \\
        & & 0-10 & 11-20 & all & 0-15 & 16-20 & all & 0-15 & 16-20 & all & 0-19 & 20 & all \\
        \midrule 
        Joint & Res101 & 78.8 & 77.7 & 78.3 & 79.8 & 73.4 & 78.3 & 79.8 & 73.4 & 78.3 & 78.2 & 80.0 & 78.3 \\
        \midrule 
        ILT~\cite{ilt} & Res101 & 7.2 & 3.7 & 5.5 & 9.6 & 7.8 & 9.2 & 67.8 & 40.6 & 61.3 & 68.2 & 12.3 & 65.5 \\ 
        SDR~\cite{sdr} & Res101 & 32.4 & 17.1 & 25.1 & 44.7 & 21.8 & 39.2 & 75.4 & 52.6 & 69.9 & 69.1 & 32.6 & 67.4 \\
        PLOP+UCD~\cite{ucd} & Res101 & 42.3 & 28.3 & 35.3 & 66.3 & 21.6 & 55.1 & 75.0 & 51.8 & 69.2 & 75.9 & 39.5 & 74.0 \\ 
        SPPA~\cite{sppa} & Res101 & - & - & - & 66.2 & 23.3 & 56.0 & 78.1 & 52.9 & 72.1 & 76.5 & 36.2 & 74.6 \\ 
        MiB+AWT~\cite{awt}  & Res101 & 33.2 & 18.0 & 26.0  & 59.1 & 17.2 & 49.1 & 77.3 & 52.9 & 71.5 & - & - & - \\ 
        ALIFE~\cite{alife} & Res101 & - & - & - & 64.4 & 34.9 & 57.4 & 77.2 & 52.5 & 71.3 & 76.6 & 49.4 & 75.3 \\ 
        GSC~\cite{cong2023gradient} & Res101 & 50.6 & 17.3 & 34.7 & 72.1 & 24.4 & 60.8 & 78.3 & 54.2 & 72.6 & 76.9 & 42.7 & 75.3 \\ 
        \midrule
        MiB~\cite{mib} & Res101 & 12.2 & 13.1 & 12.6 & 38.0 & 13.5 & 32.2 & 76.4 & 49.4 & 70.0 & 71.2 & 22.1 & 68.9 \\ 
        MiB*~\cite{mib} & Res101 & 10.4 & 9.9 & 10.2 & 45.2 & 15.7 & 38.2 & 76.8 & 49.1 & 70.2 & 71.6 & 28.6 & 69.6 \\ 
        \rowcolor{LightBlue} MiB+NeST~(Ours)  & Res101 & 52.3 & 21.0 & 37.4 & 61.7 & 20.4 & 51.8 & 77.1 & 50.1 & 70.7 & 71.7 & 28.2 & 69.7 \\ 
        \midrule 
        PLOP~\cite{plop} & Res101 & 44.0 & 15.5 & 30.5 & 65.1 & 21.1 & 54.6 & 75.7 & 51.7 & 70.1 & 75.4 & 37.4 & 73.5 \\ 
        PLOP*~\cite{plop} & Res101 & 45.9 & 17.1 & 32.2 & 66.8 & 22.3 & 56.2 & 77.0 & 50.9 & 70.8 & 75.7 & 39.4 & 74.0 \\ 
        \rowcolor{LightBlue} PLOP+NeST~(Ours)  & Res101 & 54.2 & 17.8 & 36.9 & 72.2 & 33.7 & 63.1 & 77.6 & 55.8 & 72.4 & 77.0 & 49.1 & 75.7 \\ 
        \midrule 
        RCIL~\cite{rcil} & Res101 & 55.4 & 15.1 & 34.3 & 70.6 & 23.7 & 59.4 & 78.8 & 52.0 & 72.4 & 77.0 & 31.5 & 74.7 \\ 
        RCIL*~\cite{rcil} & Res101 & 47.8 & 17.0 & 33.1 & 69.9 & 23.9 & 58.9 & 78.8 & 52.4 & 72.5 & 76.8 & 28.9 & 74.5 \\ 
        \rowcolor{LightBlue} RCIL+NeST~(Ours) & Res101 & 51.4 & 20.9 & 36.8 & 71.9 & 28.0 & 61.4 & 79.0 & 52.8 & 72.8 & 77.0 & 33.3 & 74.9 \\ 
        \midrule 
        Joint & Swin-B & 80.4 & 79.7 & 80.1 & 81.1 & 76.7 & 80.1 & 81.1 & 76.7 & 80.1 & 80.0 & 80.7 & 80.1 \\
        \midrule
        MiB*~\cite{mib} & Swin-B & 11.4 & 18.9 & 15.0 & 35.0 & 43.2 & 36.9 & 80.7 & 66.5 & 77.3 & 79.2 & 60.2 & 78.3 \\  
        \rowcolor{LightBlue} MiB+NeST~(Ours) & Swin-B & 65.2 & 35.8 & 51.2 & 77.0 & 53.3 & 71.4 & 81.2 & 67.4 & 77.9 &79.7 & 60.0 & 78.8 \\ 
        \midrule
        PLOP*~\cite{plop} & Swin-B & 37.8 & 23.1 & 30.8 & 74.1 & 52.1 & 68.9 & 80.1 & 68.1 & 77.2 & 77.0 & 65.8 & 76.4 \\ 
        \rowcolor{LightBlue} PLOP+NeST~(Ours)  & Swin-B & 64.3 & 28.3 & 47.2 & 76.8 & 57.2 & 72.2 & 80.5 & 70.8 & 78.2 & 79.6 & 70.2 & 79.1 \\
        \bottomrule
    \end{tabular}
    }
\end{table}

\subsection{Results}\label{sec:results}
In this section, we apply NeST to three classic methods, 
MiB~\cite{mib}, PLOP~\cite{plop} and RCIL~\cite{rcil} with different backbones.

\begin{figure*}[!t] 
	\centering
	\begin{small}
		\centering
		\begin{subfigure}{0.42\linewidth} 
			\includegraphics[width=\linewidth]{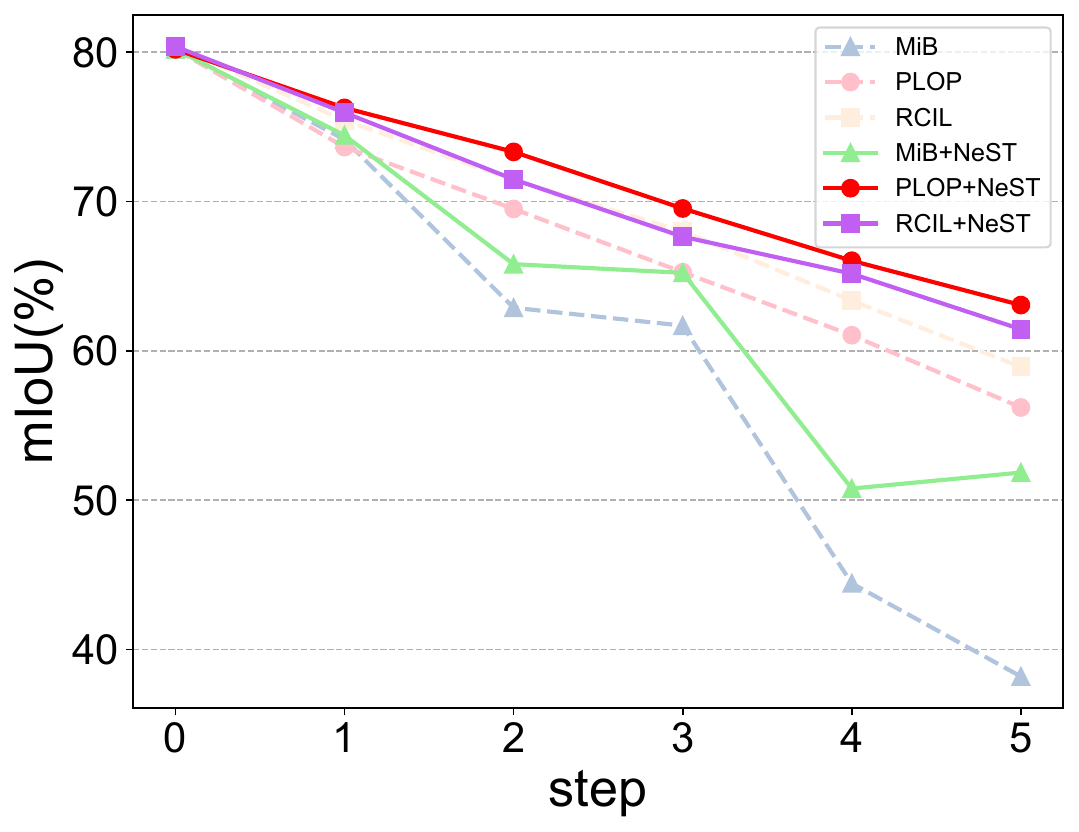}
			\caption{\emph{15-1} on PASCAL VOC 2012}
		\end{subfigure}
		\begin{subfigure}{0.42\linewidth}
            \includegraphics[width=\linewidth]{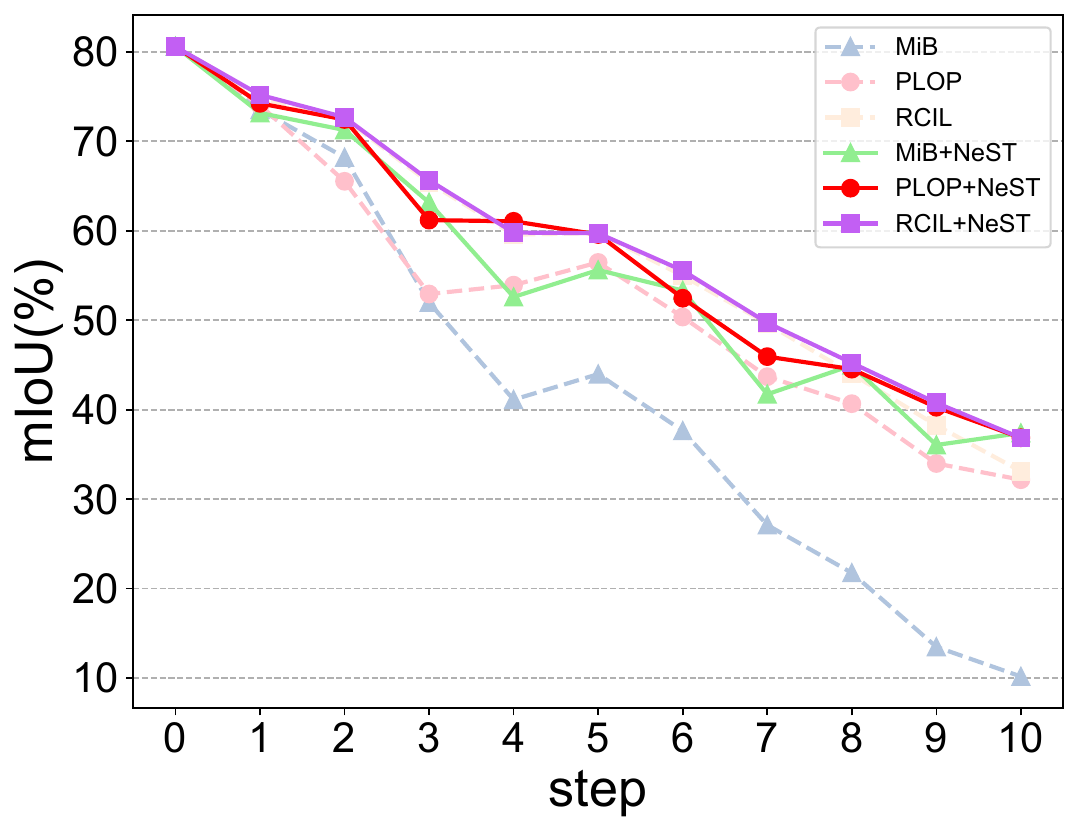}
			\caption{\emph{10-1} on PASCAL VOC 2012}
		\end{subfigure}
	\end{small}
	\caption{ The mIoU~(\%) at each step for the setting \emph{15-1}~(a) and \emph{10-1}~(b).
	}\label{fig:each_step_performance}
\end{figure*}

\minisection{Pascal VOC 2012.} We conduct experiments on \emph{10-1}, \emph{15-1}, \emph{15-5} and \emph{19-1} settings.
As shown in~\cref{tab:voc}, with NeST, MiB~\cite{mib}, PLOP~\cite{plop} and RCIL~\cite{rcil} can achieve huge performance gains.
On the \emph{15-1} setting with Deeplab, 
by simply using the proposed new classifier strategy,
we can improve the performance of MiB, PLOP, and RCIL by 13.6\%, 6.9\%, and 2.5\%, respectively.
In the more difficult \emph{10-1} scenario, NeST can boost the performance of MiB by a large margin, achieving an improvement of 27.2\%.
Meanwhile, results with Swin-B show that NeST is also suitable for Transformer-based models. 
On MiB with Swin-B, NeST can improve the performance by 34.5\% on the \emph{15-1} setting and 36.2\% on the \emph{10-1} setting.
In~\cref{tab:voc}, 
the results across all settings for old and new classes demonstrate that NeST significantly enhances stability by aligning new classifiers with the existing backbone through the pre-tuning process, thereby mitigating the forgetting of old knowledge.
We also report the mIoU of our method and \emph{baselines} at each step.
As shown in~\cref{fig:each_step_performance}, The performance of our method surpasses \emph{baselines'} performance during the whole training process.

\begin{table}[!t]\label{ade}
    \caption{The mIoU~(\%) of the last step on the ADE20K dataset for four different CISS
scenarios. *~implies results from the re-implementation of the official code.}\label{tab:ade}
    \centering
    \resizebox{1.0\textwidth}{!}{
    \begin{tabular}{c|c|ccc|ccc|ccc|ccc}
        \toprule
        \multirow{2}{*}{\textbf{Method}} & \multirow{2}{*}{\textbf{Backbone}} & \multicolumn{3}{c|}{\textbf{100-50(2 steps)}} & \multicolumn{3}{c|}{\textbf{100-10(6 steps)}} & \multicolumn{3}{c|}{\textbf{100-5(11 steps)}} & \multicolumn{3}{c}{\textbf{50-50(3 steps)}} \\
        & & 0-100 & 101-150 & all & 0-100 & 101-150 & all & 0-100 & 101-150 & all & 0-50 & 51-150 & all \\
        \midrule 
        Joint~\cite{rcil}  & Res101 & 44.3 & 28.2 & 38.9 & 44.3 & 28.2 & 38.9 & 44.3 & 28.2 & 38.9 & 51.1 & 33.3 & 38.9 \\
        \midrule 
        ILT~\cite{ilt}  & Res101 & 18.3 & 14.8 & 17.0  & 0.1 & 2.9 & 1.1 & 0.1 & 1.3 & 0.5 & 13.6 & 6.2 & 9.7 \\
        PLOP+UCD~\cite{ucd} & Res101 & 42.1 & 15.8 & 33.3 & 40.8 & 15.2 & 32.3 & - & - & - & 47.1 & 24.1 & 31.8 \\ 
        SPPA~\cite{sppa} & Res101 & 42.9 & 19.9 & 35.2 & 41.0 & 12.5 & 31.5 & - & - & - & 49.8 & 23.9 & 32.5 \\ 
        MiB+AWT~\cite{awt} & Res101 & 40.9 & 24.7 & 35.6 & 39.1 & 21.3 & 33.2 & 38.6 & 16.0 & 31.1 & 46.6 & 26.9 & 33.5 \\ 
        ALIFE~\cite{alife} & Res101 & 42.2 & 23.1 & 35.9 & 41.0 & 22.8 & 35.0 & - & - & - & 49.0 & 25.7 & 33.6 \\ 
        GSC~\cite{cong2023gradient} & Res101 & 42.4 & 19.2 & 34.8 & 40.8 & 17.6 & 32.6 & 39.5 & 11.2 & 30.2 & 46.2 & 26.2 & 33.0 \\ 
        \midrule 
        MiB~\cite{mib} & Res101 & 40.5 & 17.2 & 32.8 & 38.2 & 11.1 & 29.2 & 36.0 & 5.6 & 25.9 & 45.6 & 21.0 & 29.3 \\ 
        MiB*~\cite{mib} & Res101 & 40.5 & 23.5 & 34.9 & 37.8 & 12.1 & 29.3 & 35.8 & 6.0 & 25.9 & 45.9 & 23.9 & 31.3 \\ 
        \rowcolor{LightBlue} MiB+NeST~(Ours)  & Res101 & 40.3 & 24.6 & 35.1 & 40.2 & 20.6 & 33.7 & 39.9 & 18.0 & 32.7 & 45.6 & 26.8 & 33.2 \\ 
        \midrule 
        PLOP~\cite{plop} & Res101 & 41.9 & 14.9 & 32.9  & 40.5 & 13.6 & 31.6 & 39.1 & 7.8 & 28.7 & 48.8 & 21.0 & 30.4 \\ 
        PLOP*~\cite{plop} & Res101 & 42.2 & 15.3 & 33.3 & 41.0 & 13.7 & 32.0 & 39.6 & 8.2 & 29.2 & 48.5 & 20.8 & 30.2 \\ 
        \rowcolor{LightBlue} PLOP+NeST~(Ours)  & Res101 & 42.2 & 24.3 & 36.3 & 40.9 & 22.0 & 34.7 & 39.3 & 17.4 & 32.0  & 48.7 & 27.7 & 34.8 \\
        \midrule
        RCIL~\cite{rcil} & Res101 & 42.3 & 18.8 & 34.5 & 39.3 & 17.6 & 32.1 & 38.5 & 11.5 & 29.6 & 48.3 & 25.0 & 32.5 \\ 
        RCIL*~\cite{rcil} & Res101 & 42.3 & 16.5 & 33.8 & 39.9 & 15.7 & 31.9 & 39.1 & 12.2 & 30.2 & 48.2 & 23.6 & 31.9 \\ 
        \rowcolor{LightBlue} RCIL+NeST~(Ours) & Res101 & 42.3 & 22.8 & 35.8 & 40.7 & 19.0 & 33.5 & 39.4 & 15.5 & 31.5 & 48.2 & 27.4 & 34.4\\ 
        \midrule
        Joint & Swin-B & 43.4 & 31.9 & 39.6 & 43.4 & 31.9 & 39.6 & 43.4 & 31.9 & 39.6 & 50.7 & 33.9 & 39.6 \\
        \midrule
        MiB* & Swin-B & 42.7 & 26.1 & 37.2 & 40.2 & 15.0 & 31.8 & 39.1 & 8.6 & 29.0 & 48.3 & 26.8 & 34.1 \\ 
        \rowcolor{LightBlue} MiB+NeST~(Ours)  & Swin-B & 42.8 & 27.8 & 37.9 & 41.8 & 23.8 & 35.9 & 40.5 & 19.9 & 33.7 & 49.7 & 29.3 & 36.2 \\ 
        \midrule
        PLOP* & Swin-B & 43.4 & 17.1 & 34.7 & 41.4 & 17.7 & 33.6 & 39.7 & 13.6 & 31.0  & 50.5 & 24.1 & 33.0 \\ 
        \rowcolor{LightBlue} PLOP+NeST~(Ours)  & Swin-B & 43.5 & 26.5 & 37.9 & 41.7 & 24.2 & 35.9 & 39.7 & 18.3 & 32.6 & 50.6 & 28.9 & 36.2 \\ 
        \bottomrule
    \end{tabular}
}
\end{table}

\minisection{ADE20K.} To further evaluate NeST, we conduct experiments on the more challenging settings of the ADE20K dataset.
Results of \emph{100-50}, \emph{100-10}, \emph{100-5}and \emph{50-50} settings are shown in~\cref{tab:ade}.
NeST outperforms other competing methods.
On the most difficult \emph{100-5} setting, NeST improves MiB~\cite{mib}, PLOP~\cite{plop} and RCIL~\cite{rcil} by 6.8\%, 2.8\% and 1.3\%, which indicates that NeST is also applicable to scenarios with a large number of classes.

\begin{figure}[!t] 
	\centering
	
		\centering
		\begin{subfigure}{0.42\linewidth}
			\includegraphics[width=\linewidth]{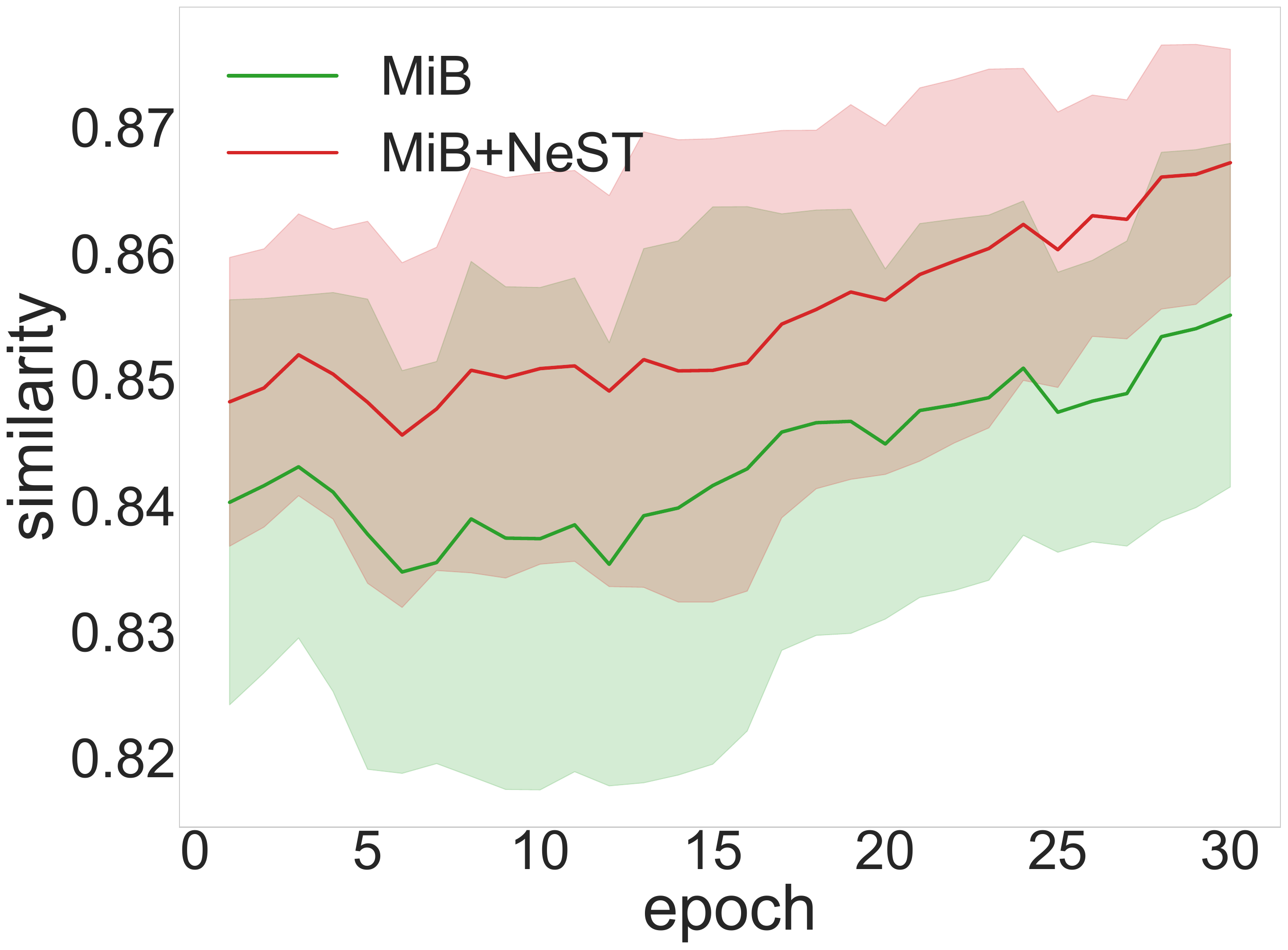}
			\caption{feature map similarity}
            \label{fig:feature_sim}
		\end{subfigure}
		\begin{subfigure}{0.42\linewidth}
			\includegraphics[width=\linewidth]{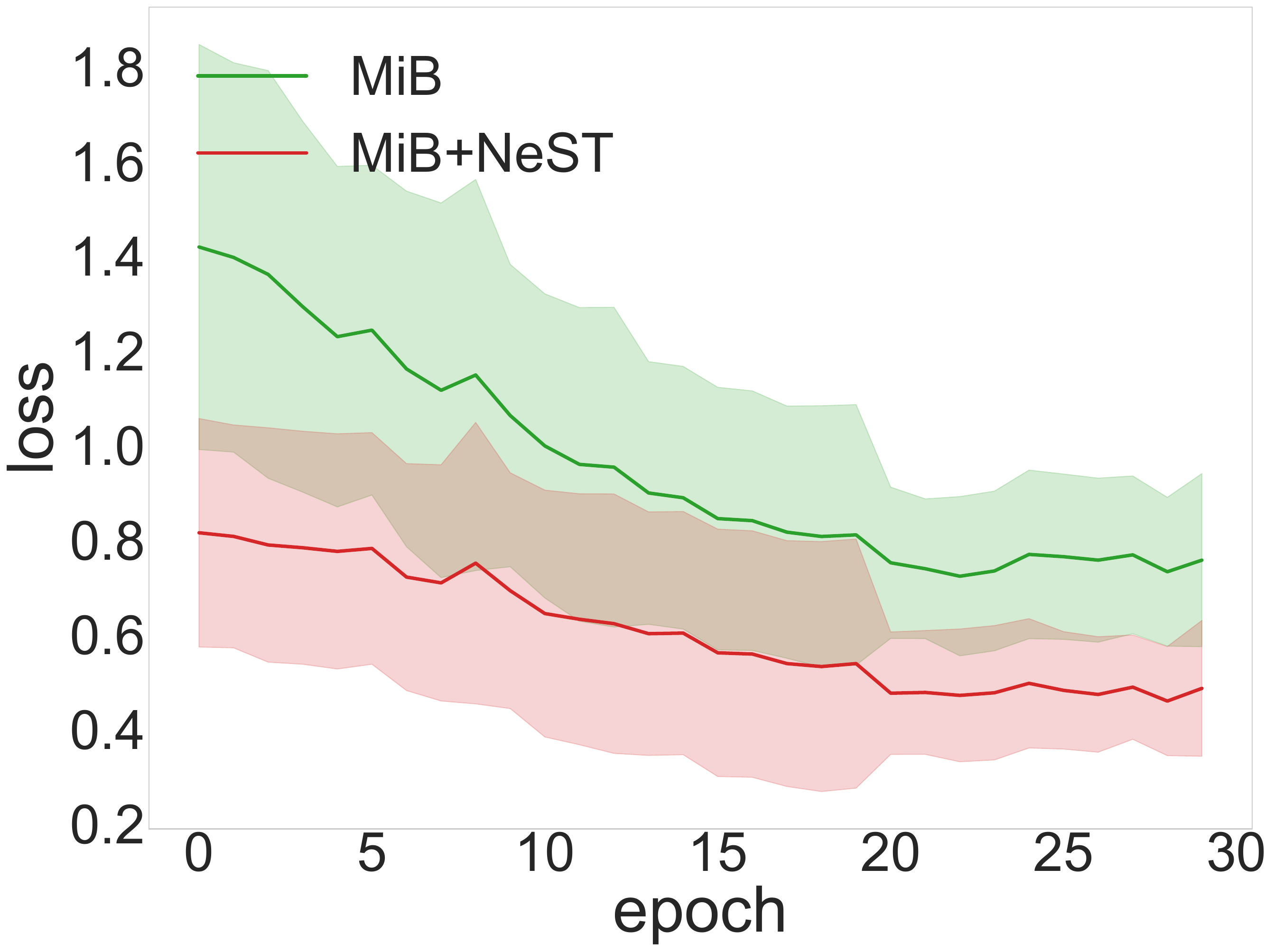}
			\caption{training loss}
   \label{fig:training_loss}
		\end{subfigure}
	
	\caption{The feature map similarity and training loss on Pascal VOC 2012 \emph{15-1} overlapped setting.
	}\label{fig:stability_gap}

\end{figure}

\minisection{Effectiveness of Our Method.} 
It is crucial for pre-tuning new classifiers before the formal training step, not only for learning new classes but also for bridging the \emph{stability gap}~\cite{de2022continual},
which means that the model will undergo a short period of forgetting when the task changes and then gradually recover.
This phenomenon is also discovered in CISS scenarios~\cite{dkd, ewf}.
At the beginning of each formal training step, a well-tuned new classifier can make the loss smaller, thus there will also be less impact on other model parameters as the gradient values become smaller. 
As the model's parameters are inherited from the old model, the old knowledge learned from previous steps can be preserved.
We conduct experiments on the \emph{15-1} setting with the \emph{baseline} MiB~\cite{mib} and MiB+NeST.
For a fair and plain comparison, we plot losses and cosine feature similarities by the mean and standard deviation of each epoch at the formal training process of the second step, while in the first step, we use the original MiB for all experiments. 
As shown in~\cref{fig:feature_sim}, 
the feature similarity of NeST exceeds that of MiB throughout the entire recovery process, demonstrating the ability of NeST to help bridge the stability gap in the model.
\cref{fig:training_loss} shows that during the formal training process, the loss of NeST is lower than the loss of MiB, helping the model converge faster and learn better.

\subsection{Ablation Study}\label{sec:ablation}

\minisection{Different Classifier Initialization Methods.}
We compare NeST with other initialization ways.
As shown in~\cref{tab:ablation_different_init}, random initialization gets the worst performance because of the misalignment in the early stages of training.
Initialization by the background classifier can get better performance.
Directly tuning the weight of new classifiers initialized by the background classifier before the formal training step 
can bring a slight increase in performance.
Unlike the aforementioned approaches, NeST enables new classifiers to adapt to training data and enhances the performance of the \emph{baseline} by 16.5\% on old classes and 4.7\% on new classes.

\begin{table}[!h]
\caption{Ablation study of different classifier initialization strategies. The performance is reported on Pascal VOC 2012 15-1 overlapped setting with MiB~\cite{mib}.}\label{tab:ablation_different_init}
\centering
\small
\setlength{\tabcolsep}{6pt}
\begin{tabular}{l|c|c|c}
\toprule
\textbf{Method}   & \textit{0-15} & \textit{16-20} & \textit{all} \\ \hline
Random   & 43.5  & 4.2 & 34.1 \\
Background~\cite{mib}   & 45.2 & 15.7 & 38.2 \\ 
Two-Stage & 46.0 & 15.3 & 38.7 \\
NeST~(Ours) & 61.7 & 20.4 & 51.8 \\
\bottomrule
\end{tabular}
\end{table}

\begin{table}[!h]
\caption{Comparison between different transformation matrix initialization methods. The performance is reported on Pascal VOC 2012 15-1 overlapped setting. }\label{tab:ablation_P_and_M}
\centering
\begin{tabular}{c|c|c|c}
\toprule
\textbf{Method}   & \textit{0-15} & \textit{16-20} & \textit{all} \\ \hline
Random Matrix Initialization   & 53.8  & 7.5 & 42.8 \\
NeST~(Ours) & 61.7 & 20.4 & 51.8 \\
\bottomrule
\end{tabular}
\end{table}

\minisection{Transformation Matrices Initialization.}
According to~\cref{tab:ablation_P_and_M}, if we use random initialization instead of our designed strategy, the overall performance will still be higher than the performance of \emph{baseline}~\cite{mib}.
However, the mIoU of new classes learned in incremental steps is significantly lower than \emph{baseline}'s performance. 
It means that if we random initialize importance matrices and projection matrices, it will only help to maintain the stability of the model.
The second row in~\cref{tab:ablation_P_and_M} shows that after initializing importance matrices and projection matrices 
using our proposed strategy increases the performance of new classes to 20.4\%, while also improving the performance of the old classes.
The design of initial values for matrices used in the transformation can facilitate the model's learning of new classes.
Thus, by initializing matrices with our strategy, NeST can achieve the trade-off between stability and plasticity.

\begin{figure*}[!htp]
	\small
	\centering
	\includegraphics[width=0.98\linewidth]{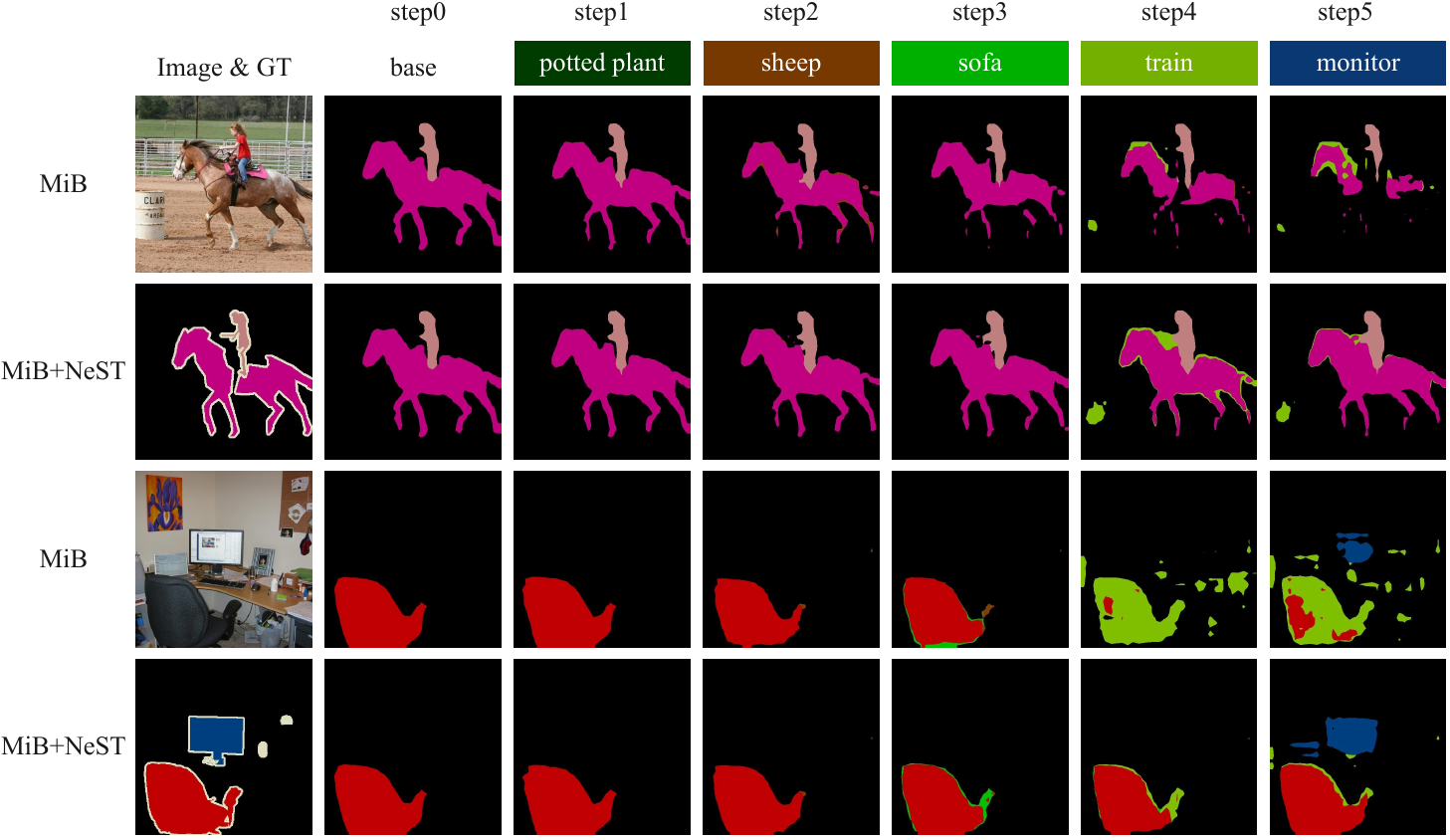}
	\caption{ Qualitative comparisons on Pascal VOC 2012 \emph{15-1} overlapped setting.
	}\label{fig:visualize}
\end{figure*}

\minisection{Component-Level Ablation Study.}
We conduct a component-level ablation study on two kinds of matrices, as shown in~\cref{tab:component_level}.
Eliminating the importance matrix means that we set values in the importance matrix to 1 and freeze it, \ie, only learning a combination of old classifiers to get new classifiers. Similarly, eliminating the projection matrix indicates that we average the weighted old classifiers to generate new classifiers. Experimental results demonstrate that with equally treated channels, the performance dramatically drops due to the unawareness of different channels.

\begin{table}[!t]
	\centering
	\small

        \caption{Component level ablation study on Pascal VOC 2012 \emph{15-1} overlapped setting.}\label{tab:component_level}
	\begin{tabular}{ccc|c|c|c}
	\toprule
	   Method & Projection Matrix & Importance Matrix & \emph{0-15} & \emph{16-20} & \emph{all}   \\ \hline
            \emph{baseline} & & & 45.2 & 15.7 & 38.2 \\
	   Variants & \checkmark & & 53.3 & 15.4 & 44.3 \\ 
          & & \checkmark & 59.7 & 19.2 & 50.1 \\

        & \checkmark & \checkmark & \textbf{61.7} & \textbf{20.4} & \textbf{51.8} \\

	\bottomrule
	\end{tabular}

  \end{table}

\minisection{Computational Costs.}
The number of extra parameters can be represented as $n_{new}\times n_{old}\times (d+1)+d+n_{new}+1$ and $n_{old}$ increases linearly with steps while $n_{new}$ is always fixed.
The additional FLOPs~(about 15.6K) and parameters~(about 5.4K) are negligible during the pre-tuning process of Step 5 on the \emph{15-1} setting.
The whole training process of MiB+NeST takes an additional 2.7\%~(3.7 minutes) of the total training time~(2.3 hours).
Note that after pre-tuning, extra parameters are removed and the formal training process is the same as MiB.

\minisection{Discussion on AWT v.s. NeST.}
AWT~\cite{awt} utilizes the new training data and old background classifier to generate new classifiers. Here we discuss the differences between AWT and NeST. AWT employs gradient-based attribution to transfer the most relevant weights to new classifiers. However, it treats each new class equally, neglecting the role of other old classifiers and the differences between new classes. Meanwhile, the gradient-based attribution technique introduces a huge memory cost, AWT can not run on RTX 3090 even if the batch size is set to 1. NeST is based on learning, which can better make the generated classifiers' weight align with the backbone. Moreover, as shown in~\cref{tab:comp_awt}, the memory cost is acceptable, and NeST is a little faster than AWT. 
\begin{table}[!t]
    \caption{The memory costs and additional training time on Pascal VOC 2012 \emph{15-1} overlapped setting.
    }\label{tab:comp_awt}
    \centering
    \begin{tabular}{c|c|c}
    \toprule
        Method & Memory per GPU & Additional Time \\\hline 
        MiB+AWT  & $>24{\rm GB}$ & $6\%$ \\
        MiB+NeST~(Ours) & $\boldsymbol{6.47{\rm GB}}$ & $\boldsymbol{2.7\%}$ \\
    \bottomrule
    \end{tabular}
\end{table}

\minisection{Robustness of Different Class Orders.}
In CISS scenarios, the class order is crucial as the learning process of a class may boost or damage the performance of another class.
To validate the robustness of NeST, we conduct experiments on the \emph{15-1} setting with five different class orders.
As shown in~\cref{tab:class_order}, the performance of NeST is higher than the performance of \emph{baseline}.

\begin{table}[!t]
\caption{The average performance of five different class orders on Pascal VOC 2012 \emph{15-1} overlapped setting.
}\label{tab:class_order}
\centering
\begin{tabular}{c|ccccc|c}
\toprule
\textbf{Method}   & \textit{A} & \textit{B} & \textit{C} & \textit{D} & \textit{E} & avg\\ \hline
MiB & 38.2 & 28.2 & 38.5  & 38.0 & 52.0 & 39.0\\
MiB+NeST~(Ours)& 51.8 & 43.4 & 50.4 & 60.0 & 59.8 & \textbf{51.5}\\
\bottomrule
\end{tabular}
\end{table}

\subsection{Visualization}

In~\cref{fig:visualize}, we show the visualization results of our proposed method based on MiB~\cite{mib}. 
Samples are selected from step 0 and from step 5 to show the stability and plasticity of our NeST.
In the first two rows, classes \emph{person} and \emph{horse} are learned in step 0, then in the following steps, the \emph{baseline} MiB~\cite{mib} gradually forget concepts learned in step 0, while NeST helps the model to preserve old knowledge.
In the last two rows, class \emph{chair} is learned in step 0 and class \emph{tv/monitor} is learned in step 5.
After learning class \emph{train}, MiB has almost forgotten the knowledge of class \emph{chair}, while NeST can give correct predictions for most of the pixels.
In the last step, NeST can help the model learn new class \emph{tv/monitor} better than MiB, showing the plasticity of our method.

\section{Conclusions}\label{sec:conclusions}
In this work, we propose a simple yet effective new classifier pre-tuning method that can enhance the CISS ability by learning a transformation from old classifiers to new classifiers.
We further find a way to initialize matrices by utilizing the information of cross-task class similarities between old classes and new classes, helping the model achieve the stability-plasticity trade-off.
Experiments on two datasets show that NeST can significantly improve the performance of \emph{baselines}, and it can be easily applied to other CISS methods.
While our approach can lead to huge performance gains, the limitation is that it introduces additional computational overhead.
In future work, we will try our method in other continual learning tasks.

\section*{Acknowledgments}
This work is funded by  
NSFC (NO. 62206135), Key Program for International Cooperation of Ministry of Science and Technology, China (NO. 2024YFE0100700), Young Elite Scientists Sponsorship Program by CAST (NO. 2023QNRC001), Tianjin Natural Science Foundation (NO. 23JCQNJC01470), and the Fundamental Research Funds for the Central Universities (Nankai University). Computation is supported by the Supercomputing Center of Nankai University.

\bibliographystyle{splncs04}
\bibliography{main}

\clearpage
\appendix




\section{Baseline Details}
\label{sec:baseline_detail}
In this section, we introduce the \emph{baselines} used in experiments. 

\minisection{MiB.} 
In MiB~\cite{mib}, two kinds of loss are used to model the background, \ie, $\mathcal{L}_{unce}$ and $\mathcal{L}_{unkd}$:

\begin{equation}
\begin{aligned}
&\mathcal{L}_{unce}=-\frac{1}{\left| \mathcal{I} \right|} \sum_{i\in\mathcal{I}}\log \tilde{q}_x^t(i, y_i)\,, \\
&\mathcal{L}_{unkd}=-\frac{1}{\left |\mathcal{I}\right |}\sum_{i\in \mathcal{I}}\sum_{c\in \bigcup_{j=1}^{t-1}\mathcal{C}_j}q_{x}^{t-1}(i, c)\log \hat{q}_x^t(i, c)\,,
\end{aligned}
    \label{eqn:unce_supp}
\end{equation}
where $\mathcal{I}$ denotes the pixel set of an image, $y_i\in c_{bg}\cup\mathcal{C}_t$ denotes the ground-truth label of pixel $i$, $q_x^t$ denotes the output of the model at step $t$, $\tilde{q}_x^t$ and $\hat{q}_x^t$ denotes the modified output of the current model, considering the old classes for the cross entropy loss and new classes for the knowledge distillation loss.

\minisection{PLOP.}
Different from MiB~\cite{mib}, PLOP~\cite{plop} utilizes pseudo-labeling to address the issue of background shift, as follows:
\begin{equation}
    \begin{aligned}
    \mathcal{L}_{pseudo}=-\frac{\nu}{WH}\sum_{w,h}^{W, H}\sum_{c\in\mathcal{C}_t}\tilde{S}(w,h,c)\log \hat{S}^t(w, h, c)\,,
    \end{aligned}
\end{equation}
where $\hat{S}$ denotes the prediction of the model and $\tilde{S}$ denotes the pseudo-labels generated by the old model in the previous step.

It also distills intermediate features by Local POD, as follows:
\begin{equation}
    \begin{aligned}
        \mathcal{L}_{LocalPod}=\frac{1}{L}\sum_{l=1}^L\left|\left| \Phi(f_l^t(I))-\Phi(f_l^{t-1}(I)) \right|\right|\,,
    \end{aligned}
\end{equation}
where $L$ denotes the number of layers, $\Phi$ denotes the operation of Local POD embedding extraction, $f_l^t(I)$ denotes the output feature from the layer $l$ with the input $I$.

\minisection{RCIL.}
RCIL~\cite{rcil} decouple the remembering of old knowledge and the learning of new knowledge by adding a parallel module composed of a convolution layer and a normalization layer for each $3\times 3$ convolution module. At step 0, all parameters are trainable. At the beginning of each incremental step, two branches of the old model are fused into one frozen branch to memorize the old knowledge, while the other branch is learnable. A drop path strategy is also used when fusing the outputs of two branches, which can be denoted as:
\begin{equation}
    \begin{aligned}
        x_{out} = \eta\cdot x_1 + (1-\eta)\cdot x_2\,,
    \end{aligned}
\end{equation}
where $x_{out}$ denotes the fused output, $x_1$ and $x_2$ denotes the outputs from two branches, and $\eta$ denotes a channel-wise weight vector. For training process, $\eta$ is sampled from the set $\{0, 0.5, 1\}$ and for evaluation $\eta$ is set to 0.5. RCIL also proposed a Pooled Cube Knowledge Distillation, using average pooling operation on spatial and channel dimensions.

\section{Results and Analysis of Disjoint Settings}
In \emph{Disjoint} settings, at each step, the \emph{bg} classifier will not see any future class, leading to MiB's initialization struggling in these settings, 
while our NeST leverages semantic knowledge from old classifiers to generate new classifiers for initialization, the pre-tuning process also benefits the stability of the model.
Results of NeST and \emph{baselines} on \emph{15-1 Disjoint} and \emph{10-1 Disjoint} settings are shown in~\cref{tab:disjoint}, indicating that NeST can significantly improve the performance of previous methods in \emph{Disjoint} settings.
\begin{table}[!h]
    \caption{Results of disjoint settings on Pascal VOC 2012 dataset.}\label{tab:disjoint}
    \centering
    \small
    \setlength{\tabcolsep}{6pt} 
    \begin{tabular}{c|c|c}
    \toprule
        Method & \emph{15-1 Disjoint} & \emph{10-1 Disjoint} \\\hline 
        MiB  & 38.6 & 2.0 \\
        MiB+NeST & \textbf{41.0} & \textbf{20.5}  \\\hline 
        PLOP & 40.7 & 12.6  \\
        PLOP+NeST & \textbf{52.7} & \textbf{22.4} \\
    \bottomrule
    \end{tabular}
\end{table}

\section{Experiments of COCO-Stuff 10K}

To prove the ability to apply our NeST in scenarios with more classes, we introduce another dataset for class incremental semantic segmentation, COCO-Stuff 10K, to evaluate the effectiveness of our method. COCO-Stuff 10K includes 80 thing classes and 91 stuff classes, which is a subset of the original COCO-Stuff dataset. 
We evaluate our NeST on the \emph{80-91}~(2 Steps) overlapped setting, which contains more classes than ADE20K and Pascal VOC 2012. As shown in~\cref{tab:coco}, our method can handle scenarios with more classes in one step.

\begin{table}[!h]
    \caption{Results of \emph{80-91} overlapped setting on COCO-Stuff 10K dataset.}\label{tab:coco}
    \centering
    \small
    \setlength{\tabcolsep}{6pt} 
    \begin{tabular}{c|c|c|c}
    \toprule
        Method & \emph{0-80} & \emph{81-171} & \emph{all} \\\hline 
        MiB  & 40.2 & 18.6 & 28.9 \\
        MiB+NeST & 41.9 & 20.7 & \textbf{30.7} \\\hline 
        PLOP & 46.1 & 17.0 & 30.8 \\
        PLOP+NeST & 46.0 & 18.8 & \textbf{31.6} \\
    \bottomrule
    \end{tabular}
\end{table}

\section{Comparisons with Transformer-based SOTA Methods}
In recent years, many Transformer-based CISS methods have emerged, here we briefly discuss the differences between NeST and these methods.
Comformer~\cite{cermelli2023comformer} uses a universal segmentation model Mask2Former~\cite{cheng2022masked} to do mask classification for continual panoptic segmentation and continual semantic segmentation.
CoinSeg~\cite{zhang2023coinseg} introduces a pretrained Mask2Former model as class-agnostic mask generator.
Incrementer~\cite{shang2023incrementer} sequentially adds tokens of new classes and performs dot production between features and updated class tokens to generate segmentation prediction results.
The \emph{baseline} is based on a simple per-pixel classification model SETR with ViT-B as the backbone, while equipped with NeST, it can achieve SOTA performances, as shown in~\cref{tab:comp_sota}.
Moreover, NeST has the potential to be integrated into these transformer-based methods, and we leave it as our future work.

\begin{table}[!h]
\caption{Comparisons with Transformer-base SOTA methods.}\label{tab:comp_sota}
	\centering
	\small
	\setlength\tabcolsep{1.0mm}
	\renewcommand{\arraystretch}{1.3}
	\begin{tabular}{c|c|c|c|c|c}
	\toprule
            Method & Backbone & Model & \emph{15-1} & \emph{15-5} & \emph{10-1} \\\hline
            CoinSeg & Swin-B & Deeplab+Mask2Former & 75.5 & 77.6 & 70.5 \\
            Incrementer & ViT-B & Segmenter & 75.5 & 79.9 & 70.2 \\
            MiB & ViT-B & SETR & 53.3 & 80.2 & 25.5 \\
            MiB+NeST~(Ours) & ViT-B & SETR  & \textbf{76.5} & \textbf{80.3} & \textbf{71.9} \\
	   \bottomrule
	\end{tabular}
	\label{tab:fabl}
  \end{table}

\section{More Implementation Details}
\minisection{Weight Align.} To prevent the new classifiers' weight from being too large during the pre-tuning process, we apply Weight Aligning~(WA)~\cite{weight_align} as follows:
\begin{equation}
    \begin{aligned}
        \hat{w}_{new} = w_{new} \cdot \frac{Mean(Norm_{old})}{Mean(Norm_{new})}
    \end{aligned}
\end{equation}
where $Norm_{old}$ and $Norm_{new}$ denote norms of old and new classifiers' weights, $Mean(\cdot)$ denotes the operation of calculating mean values.
Relevant experiment results in~\tabref{tab:ablation_WA} show that WA can correct the biased weight thus boosting the performance of NeST.
\begin{table}[!h]
\caption{
Ablation study of Weight Align for NeST. All performances are reported on the \emph{15-1} setting. }\label{tab:ablation_WA}
\centering
\small
\setlength{\tabcolsep}{6pt} 
\begin{tabular}{l|c|c|c}
\toprule
\textbf{Method}   & \textit{0-15} & \textit{16-20} & \textit{all} \\ \hline
MiB+NeST w/o WA & 58.4 & 10.9 & 47.1 \\ 
MiB+NeST w/ WA & 61.7 & 20.4 & \textbf{51.8} \\ \hline
PLOP+NeST w/o WA & 72.5 & 32.4 & 62.9 \\
PLOP+NeST w/ WA & 72.2 & 33.7 & \textbf{63.1} \\
\bottomrule
\end{tabular}
\end{table}

\minisection{Fix old classifiers. }
We find that the pseudo-labeling strategy may change the geometric structure of old classifiers severely, which has a detrimental impact on our method. This phenomenon is particularly obvious on the Pascal VOC 2012 dataset.
To preserve the old knowledge learned in previous steps, following EWF~\cite{ewf}, we fix old classifiers in the formal training steps on settings of Pascal VOC 2012.
Relevant experiment results are shown in~\tabref{tab:ablation_fix}.
\begin{table}[!htp]
\caption{
Ablation study of fixing previous classifiers for our method based on PLOP~\cite{plop}. All performances are reported on the \emph{15-1} setting. }\label{tab:ablation_fix}
\centering
\small
\setlength{\tabcolsep}{6pt}
\begin{tabular}{l|c|c|c}
\toprule
\textbf{Method}   & \textit{0-15} & \textit{16-20} & \textit{all} \\ \hline
PLOP w/ fix & 56.9 & 11.3 & 46.0 \\
PLOP+NeST w/o fix & 66.8 & 20.2 & 55.7 \\ 
PLOP+NeST w/ fix & 72.2 & 33.7 & \textbf{63.1} \\
\bottomrule
\end{tabular}
\end{table}

\section{Further Analysis}

\begin{figure*}[!htp]
	\small
	\centering
	\includegraphics[width=0.89\linewidth]{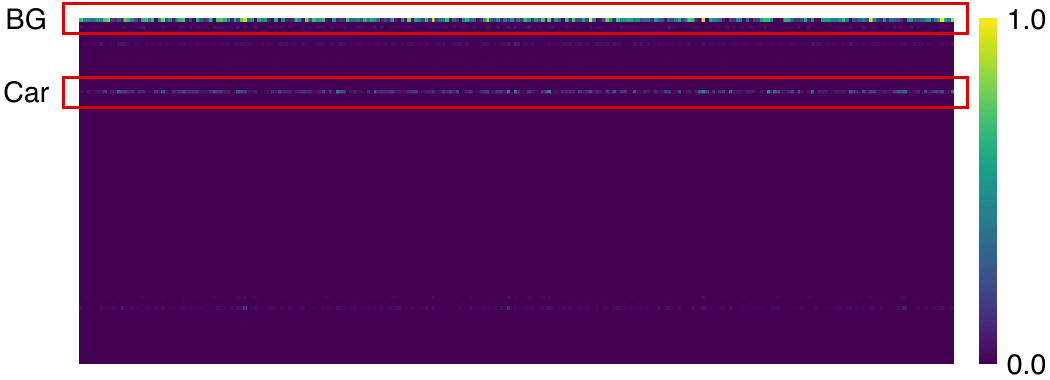}
	\caption{ Visualization of the importance matrix on ADE20K \emph{100-5} step1.
	}\label{fig:weight_old_viz}
\end{figure*}
\minisection{Effectiveness of the importance matrix.}
For plasticity, we learn to generate a new classifier with relevant old classifiers.
In particular, the importance matrix can capture the semantic relationship between old and new classifiers on the channel level. To verify this, we visualize the importance matrix $M\in\mathbb{R}^{100\times 256}$ of the new class \emph{van} on ADE20K \emph{100-5} step1. We normalized the absolute values to [0, 1] and a lighter color means a higher value. As shown in~\cref{fig:weight_old_viz}, the bg class (row: 0) and car class (row: 21) make the largest contributions. It is intuitive, as the class van may appear in old data, labeled as bg, and van and old class car are closer in semantic relationship.

\begin{figure*}[!h]
	\small
	\centering
	\includegraphics[width=0.99\linewidth]{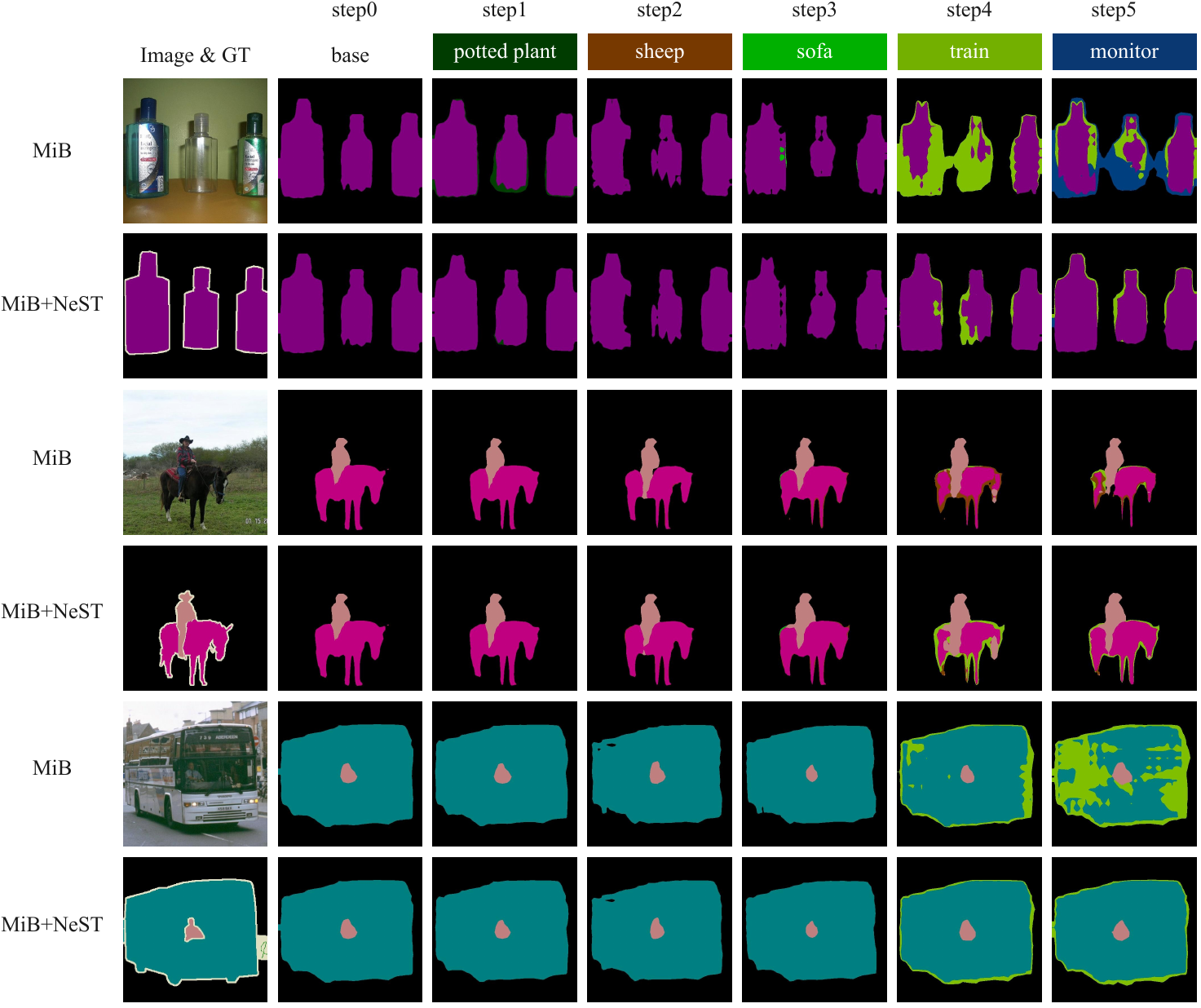}
	\caption{ More qualitative results. All experiments are conducted on the \emph{15-1} setting.
	}\label{fig:more_qua_1}
\end{figure*}

\begin{figure*}[!h]
	\small
	\centering
	\includegraphics[width=0.99\linewidth]{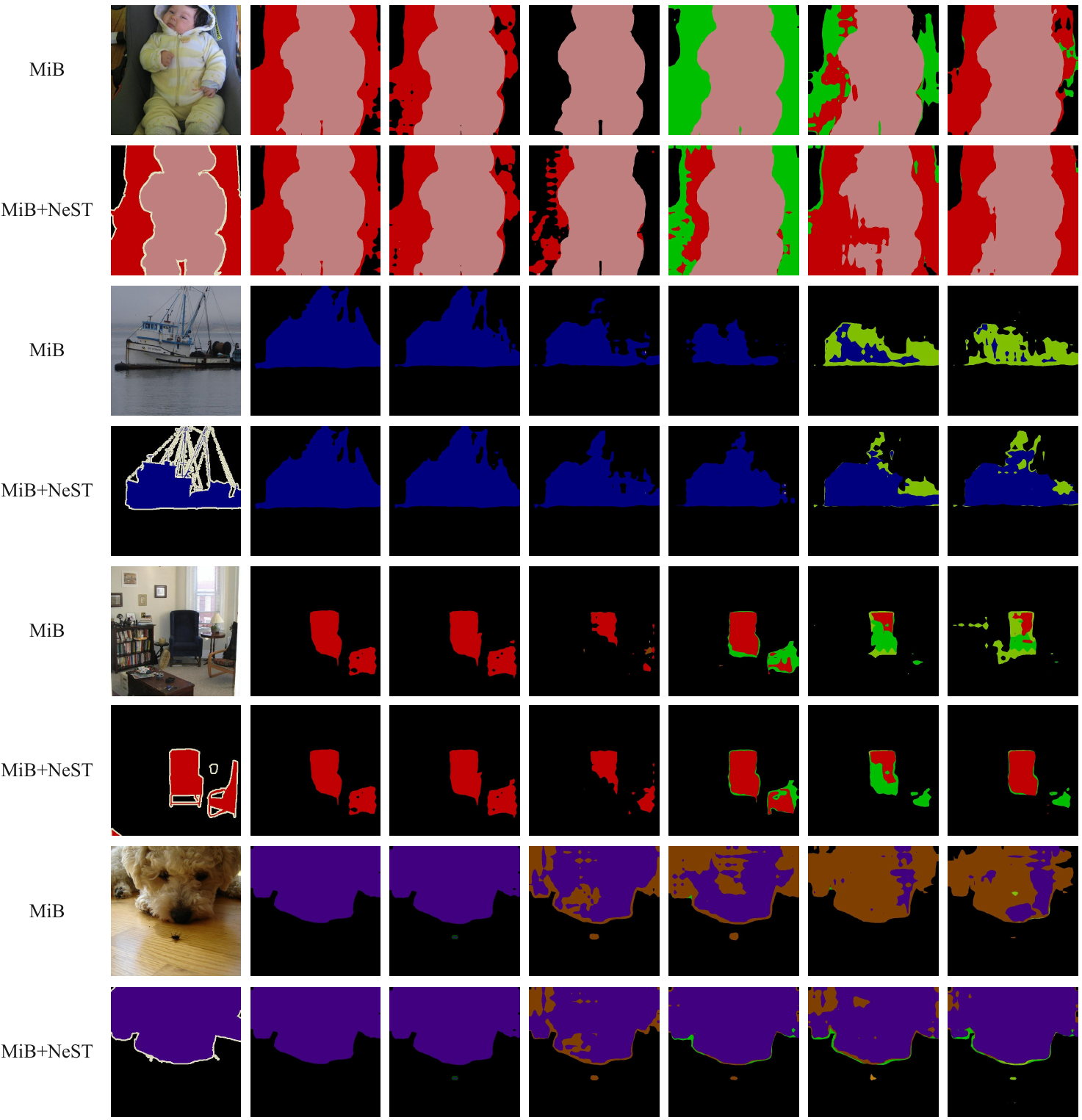}
	\caption{ More qualitative results. All experiments are conducted on the \emph{15-1} setting.
	}\label{fig:more_qua_2}
\end{figure*}
\minisection{Different class orders.}
To evaluate the effectiveness of our method, following PLOP~\cite{plop}, we use five different class orders of Pascal VOC 2012 \emph{15-1} overlapped setting, as follows:
\begin{equation}
\scriptsize
\begin{aligned}
A:[0, 1 , 2 , 3 , 4 , 5 , 6 , 7 , 8 , 9 , 10 ,
11 , 12 , 13 , 14 , 15 , 16 , 17 , 18 , 19 , 20]\,,\\
B:[0, 12, 9, 20, 7, 15, 8, 14, 16, 5, 19, 4, 1, 13, 2, 11, 17, 3, 6, 18, 10]\,,\\
C:[0, 13, 19, 15, 17, 9, 8, 5, 20, 4, 3, 10, 11, 18, 16, 7, 12, 14, 6, 1, 2]\,,\\
D:[0, 15, 3, 2, 12, 14, 18, 20, 16, 11, 1, 19, 8, 10, 7, 17, 6, 5, 13, 9, 4]\,,\\
E:[0, 7, 5, 3, 9, 13, 12, 14, 19, 10, 2, 1, 4, 16, 8, 17, 15, 18, 6, 11, 20]\,.
\end{aligned}
    \label{eqn:class_order}
\end{equation}

\minisection{More qualitative results.}
More qualitative results are shown in~\figref{fig:more_qua_1} and~\figref{fig:more_qua_2}.
By applying the pre-tuning process, our method can help the model preserve old knowledge.

\begin{figure*}[!h]
	\small
	\centering
	\includegraphics[width=0.99\linewidth]{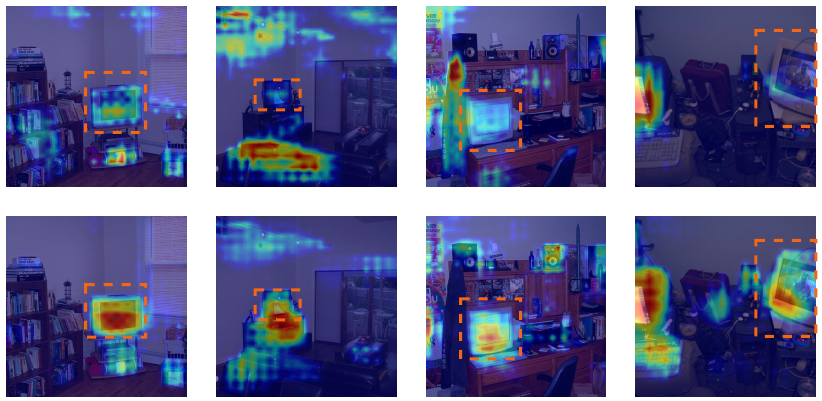}
	\caption{ Class activation maps for the last class \emph{tv/monitor} on the \emph{15-1} setting \(w \!/\! o\)~(the top row) and \(w \!/\! \)~(the bottom row) our matrix initialization strategy.
	}\label{fig:CAM}
\end{figure*}

Moreover, to validate the effectiveness of the matrix initialization, we also visualize the class activation map for the last class \emph{tv/ monitor}.
To visualize segmentation CAMs, we adopt the method proposed in~\cite{vinogradova2020towards}. As shown in~\cref{fig:CAM}, with the designed matrix initialization strategy, the model can pay more attention to areas of new classes.


%
%

\end{document}